\documentclass{article}
\PassOptionsToPackage{numbers, compress}{natbib}
\usepackage{iclr2027_conference,times}
\iclrfinalcopy

\usepackage{graphicx} \graphicspath{{figures/}}
\usepackage{amsmath,amssymb,mathabx,mathtools,amsthm,nicefrac}
\usepackage[pagebackref,breaklinks,colorlinks]{hyperref}
\usepackage[capitalise,nameinlink]{cleveref}
\usepackage[skip=3pt,font=small]{subcaption}
\usepackage[skip=3pt,font=small]{caption}
\usepackage{acronym}
\usepackage{wrapfig}
\usepackage[dvipsnames,svgnames,x11names,table]{xcolor}
\usepackage{enumitem}
\usepackage{xspace}
\usepackage[misc]{ifsym}
\usepackage{titletoc}
\usepackage{verbatim,fancyvrb,listings}
\usepackage{booktabs,tabularx,colortbl,multirow,multicol,array,makecell,tabularray}
\usepackage{siunitx}
\usepackage{bm}
\usepackage{cancel}

\crefname{section}{Sec.}{Secs.}
\Crefname{section}{Section}{Sections}
\crefname{subsection}{Sec.}{Secs.}
\Crefname{subsection}{Section}{Sections}
\crefname{subsubsection}{Sec.}{Secs.}
\Crefname{subsubsection}{Section}{Sections}
\crefname{table}{Tab.}{Tabs.}
\Crefname{table}{Table}{Tables}
\crefname{figure}{Fig.}{Figs.}
\Crefname{figure}{Figure}{Figures}
\crefname{equation}{Eq.}{Eqs.}
\Crefname{equation}{Equation}{Equations}
\crefname{appendix}{Appx.}{Appxs.}
\Crefname{appendix}{Appendix}{Appendices}

\makeatletter
\DeclareRobustCommand\onedot{\futurelet\@let@token\@onedot}
\def\@onedot{\ifx\@let@token.\else.\null\fi\xspace}

\def\vs{\emph{vs}\onedot}

\makeatother

\newcommand{\method}{DDC\xspace}
\newcommand{\CoM}{\bm{c}}
\newcommand{\dCoM}{\dot{\bm{c}}}
\newcommand{\pCoM}{\tilde{\bm{c}}}
\newcommand{\xcom}{\bm{\xi}}
\newcommand{\supc}{\bm{s}}
\newcommand{\rrel}{\bm{r}}
\newcommand{\drrel}{\dot{\bm{r}}}
\newcommand{\wnat}{\omega_0}
\newcommand{\Jc}{\bm{J}_{c}}
\newcommand{\Js}{\bm{J}_{s}}
\newcommand{\Epos}{E_{\mathrm{pos}}}
\newcommand{\Evel}{E_{\mathrm{vel}}}
\newcommand{\Eacc}{E_{\mathrm{acc}}}

\makeatletter
\renewcommand{\paragraph}{\@startsection{paragraph}{4}{\z@}{\z@}{-1em}{\normalsize\bf}}
\makeatother
\acrodef{com}[CoM]{center of mass}
\acrodef{xcom}[xCoM]{extrapolated center of mass}
\acrodef{cop}[CoP]{center of pressure}
\acrodef{mos}[MoS]{margin of stability}
\acrodef{ttb}[TTB]{time-to-boundary}
\acrodef{ttf}[TTF]{time-to-fall}
\acrodef{lip}[LIP]{linear inverted pendulum}
\acrodef{imu}[IMU]{inertial measurement unit}
\acrodef{dof}[DoF]{degree of freedom}
\acrodefplural{dof}[DoFs]{degrees of freedom}
\acrodef{rl}[RL]{reinforcement learning}
\acrodef{pd}[PD]{proportional-derivative}
\acrodef{mlp}[MLP]{multilayer perceptron}
\acrodef{rms}[RMS]{root mean square}
\acrodef{ml}[ML]{mediolateral}
\acrodef{ap}[AP]{anteroposterior}

\title{A Change of Frame Makes the Capture Point\\
    Proprioceptive: Distillation-Free Humanoid\\
    Single-Leg Balance}

\author{%
    \textbf{Yikai Zhou}$^{1,5,7}$ \quad
    \textbf{Xingyun Wang}$^{2,5,7}$ \quad
    \textbf{Jieming Cui}$^{3,4,6,8,9}$ \quad
    \textbf{Bozhou Chen}$^{1,5,7}$ \\
    \textbf{Yikai Fan}$^{2,5,7}$ \quad
    \textbf{Yixin Zhu}$^{4,3,6,8,9,10\ \textrm{\Letter}}$ \quad
    \textbf{Wenxin Li}$^{1,5,7\ \textrm{\Letter}}$ \\
    \normalfont $^{1}$School of Computer Science, Peking University \\
    \normalfont $^{2}$School of Electronics Engineering and Computer Science, Peking University \\
    \normalfont $^{3}$Institute for Artificial Intelligence, Peking University \\
    \normalfont $^{4}$School of Psychological and Cognitive Sciences, Peking University \\
    \normalfont $^{5}$National Key Laboratory of Intelligent Parallel Technology, Beijing, China \\
    \normalfont $^{6}$State Key Laboratory of General Artificial Intelligence, Peking University \\
    \normalfont $^{7}$Beijing Key Laboratory of Software and Hardware Cooperative Artificial Intelligence Systems \\
    \normalfont $^{8}$Beijing Key Laboratory of Brain-Computer Interface and Mental Health Modulation \\
    \normalfont $^{9}$Embodied Intelligence Lab, PKU-Wuhan Institute for Artificial Intelligence \\
    \normalfont $^{10}$XSVision \quad
    $^{\textrm{\Letter}}$\texttt{yixin.zhu@pku.edu.cn},
    \texttt{lwx@pku.edu.cn}\\[2pt]
    {\hypersetup{urlcolor=blue}
    \url{https://estoil.github.io/DDC/}}
}

\begin{document}

\maketitle
\pagestyle{plain}
\thispagestyle{plain}

\begin{abstract}
    Unified humanoid policies track agile whole-body motion, yet few can hold a clean single-leg stance.
    On the single-leg balance benchmark introduced here, eight released state-of-the-art general policies hold a clean stance on none of 90 held-out motions; they stay upright only by hopping or re-planting a foot, recovering from imbalance rather than preventing it.
    Prevention needs the capture point, the center of mass (CoM) extrapolated by its velocity. That velocity contains the base linear velocity, which no on-board sensor measures, so the capture point has been confined to rewards and privileged critics and has reached hardware only through teacher--student distillation.
    A change of frame removes the obstacle: expressed relative to the support foot, the base velocity cancels identically, leaving a capture-point state reconstructible from joint encoders and an inertial measurement unit alone. We place this support-relative dynamic-CoM observation directly in the deployed actor and pair it with a reward library translated term by term from human postural control.
    Trained with asymmetric FastSAC and no distillation, the resulting policy, DDC, holds clean single-leg balance on 89 of 90 held-out motions across nine pose classes and runs directly on a Unitree G1; removing the observation alone costs 43 points, and 53 under deployment noise.
    We release the policy, the data, and a method-agnostic MuJoCo benchmark for humanoid single-leg balance, which scores released policies on the same held-out motions.
    Together these turn single-leg balance from a per-task demonstration into a capability the field can measure and build into general policies.
\end{abstract}

\begin{figure}[t]
    \centering
    \includegraphics[width=\linewidth]{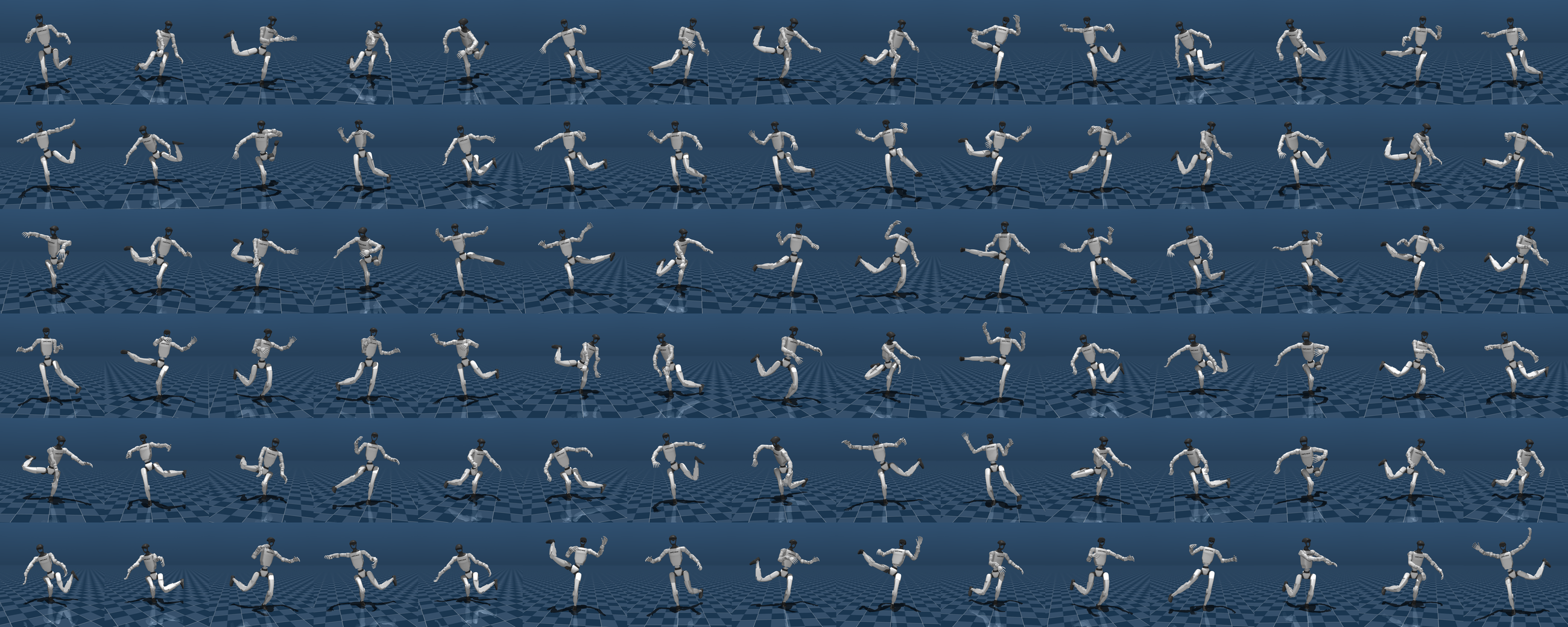}
    \caption{\textbf{Single-leg balance across the pose grid, without distillation.} \method holds stable single-leg balance on poses that vary in squat depth, swing-foot height, and support leg, and transfers directly to a real Unitree G1. We also release a method-agnostic, reproducible MuJoCo benchmark for single-leg balance that selects the checkpoint to deploy, scores any released policy under one protocol, and probes the competence of generalist policies.}
    \label{fig:teaser}
\end{figure}

\section{Introduction}\label{sec:intro}

Humanoid policies now track dances, runs, and flips from large motion datasets~\citep{exbody2024,omnih2o2024,humanplus2024,any2track2025,omnixtreme2026}, yet many cannot stand on one leg. For humans, balance underlies every motor skill, and single-leg support is its most demanding form~\citep{hof2005,riemann2003}; it is the natural stress test of whether a policy balances at all. On the benchmark we build for this test, eight state-of-the-art motion-tracking policies released between 2024 and 2026~\citep{protomotions2024,gmt2025,twist2025,sonic2025,mosaic2026,holomotion2026,humanoidgpt2026,omnixtreme2026} hold a clean single-leg stance on none of 90 held-out motions. Most do not simply fall: the strongest stay upright by re-planting the swing foot or hopping the support foot. A capable generalist is thus robust on one leg without being competent on it. It recovers from imbalance rather than preventing it.

Prevention has a well-understood signal. Whether a body can come to rest over its foot is decided by the capture point, the \ac{com} extrapolated by its velocity~\citep{hof2005,pratt2006}. Learning-based humanoid balance has used only its static half, keeping the \ac{com} position inside the support polygon~\citep{hub2025,ams2026,narrowterrain2025}, which is sound in quasi-static balance but blind to the inertial term $\dCoM/\wnat$ that dominates once motion is not. The omission is not an oversight. The \ac{com} velocity contains the base linear velocity, which no on-board sensor measures and which estimators recover only with drift and lag. The capture point has therefore stayed in training rewards and privileged critics, and reaching hardware has required distilling that privileged knowledge into a balance-blind student, as HuB~\citep{hub2025} and AMS~\citep{ams2026} both do.

We show that this barrier is an artifact of the frame. Taken relative to the support foot, the capture point loses the base velocity identically; what remains is a function of the joint encoders, the \ac{imu} gyroscope, and the kinematic model, computable on the robot with no state estimator in the loop. The deployed actor can therefore observe a support-relative dynamic-\ac{com} state, and to our knowledge this is the first time the capture point drives a learned policy on hardware. Around it we build a reward library from postural-control science, each term shaped to keep the capture point inside the foot rather than to step back under it once it has escaped.

Single-leg balance also lacks a yardstick. The specialized methods~\citep{hub2025,ams2026} release neither code nor policies and report self-chosen metrics on their own demonstrations, so their claims cannot be compared, and generalists are never tested for the skill at all. We build a method-agnostic MuJoCo benchmark: every released policy runs at the robot's command interface and is scored from its physical outcome alone, by a metric suite grounded in postural control. It yields the 0/90 finding above and selects the checkpoint we deploy.

The resulting policy, \method (Deployable Dynamic-\acs{com}), holds a clean single-leg stance on 89 of the 90 held-out motions across nine stratified pose classes and runs directly on a Unitree G1 (\cref{fig:teaser}). Ablations isolate each component: the dynamic-\ac{com} observation matters most, at $-43$ points on its own and $-53$ under deployment noise; the time-to-boundary reward is next at $-26$ (\cref{sec:ablation}). Our contributions are:
\begin{itemize}[nosep,leftmargin=*]
    \item \textbf{A proprioceptive capture point.} A support-relative dynamic-\acs{com} observation that the deployed actor computes from encoders and an \acs{imu} alone, with a derivation showing that the base velocity cancels. It yields a unified single-leg balance policy that runs on real hardware without teacher--student distillation.
    \item \textbf{A human-science reward library.} Four penalty terms translated from postural-control research and designed to prevent loss of balance rather than recover from it, each with its contribution isolated by ablation.
    \item \textbf{A method-agnostic benchmark.} A reproducible MuJoCo testbed for humanoid single-leg balance, with a stratified 900-motion set and a postural-control metric suite, on which eight released generalists all fail to hold a single clean stance.
\end{itemize}

\section{Related Work}\label{sec:related}

\paragraph{Balance in learned humanoid control.}
Rooted in physics-based motion imitation~\citep{deepmimic2018,amp2021}, \ac{rl}-based whole-body tracking in open frameworks such as BeyondMimic~\citep{beyondmimic2025} and Holosoma~\citep{holosoma2025} has driven humanoids to balance-critical skills: HuB's single-leg poses~\citep{hub2025}, KungfuBot's martial arts~\citep{kungfubot2025}, and AMS's agility with stability~\citep{ams2026}. Once the tracked motion is itself balance-critical, however, tracking accuracy no longer implies staying upright, and the bottleneck becomes how balance is represented. The quantity that governs falling is the capture point, classical in biomechanics~\citep{hof2005} and humanoid push recovery~\citep{pratt2006} and still central to model-based control~\citep{yang2025}. In learning-based humanoid balance this dynamic signal never reaches the deployed policy. Most methods use its static surrogate, rewarding the \ac{com} position for staying inside the support polygon or tracking a reference (HuB, AMS, KungfuAthleteBot~\citep{kungfuathletebot2026}, Narrow-Terrain~\citep{narrowterrain2025}), or apply it as an offline motion filter (KungfuBot). Where a genuine capture point appears, closest to us in \citet{poddar2026}, it is confined to critic inputs and reward terms with a proprioception-only actor, explicitly because \ac{com}-state estimation is hard on hardware; FAST~\citep{fast2026} is an actor-side exception that observes only a static reference \ac{com} and \ac{cop} target. Because balance is treated as unobservable on-board, HuB and AMS reach hardware only by distilling a privileged teacher into a deployable student. We remove the barrier at its root (\cref{sec:obs}).

\paragraph{Human postural control.}
Decades of postural-control research supply principles that humanoid balance rewards have largely not used: an ankle$\to$knee$\to$hip correction hierarchy~\citep{tropp1988,riemann2003}, stability read spatially as the \ac{mos}~\citep{hof2005} and temporally as the \ac{ttb}~\citep{hertel2006,mckeon_hertel2008,mckeon_hertel2008b}, and smoothness indexed by jerk~\citep{semak2020}. We translate each into a reward term and a matching metric (\cref{sec:reward}).

\paragraph{Benchmarks for humanoid control and balance.}
Three lines of benchmarks border our task. Single-simulator suites standardize humanoid skills, from loco-manipulation~\citep{humanoidbench2024} and egocentric hierarchical control~\citep{humanoidarena2026} to locomotion imitation~\citep{locomujoco2023} and simulated sports~\citep{smplolympics2024}, but each scores policies trained inside its own simulator and none isolates single-leg balance. A classical line benchmarks balance on physical apparatus, through human-inspired posture-control protocols~\citep{mergner_lippi2018}, \ac{com}-stabilization controllers~\citep{castano2022}, and push-recovery resilience~\citep{monteleone2023}, targeting model-based controllers on perturbation rigs rather than learned policies. Closest to us, Switch-JustDance~\citep{kim2025} is method-agnostic, scoring released whole-body tracking checkpoints through a console-game pipeline for full-body dance, not single-leg balance in simulation. None provides what a comparison of released single-leg balance policies needs: a testbed that is method-agnostic and reproducible in software alone (\cref{sec:benchmark}).

\section{Method}\label{sec:method}

\begin{figure}[t]
    \centering
    \includegraphics[width=\linewidth]{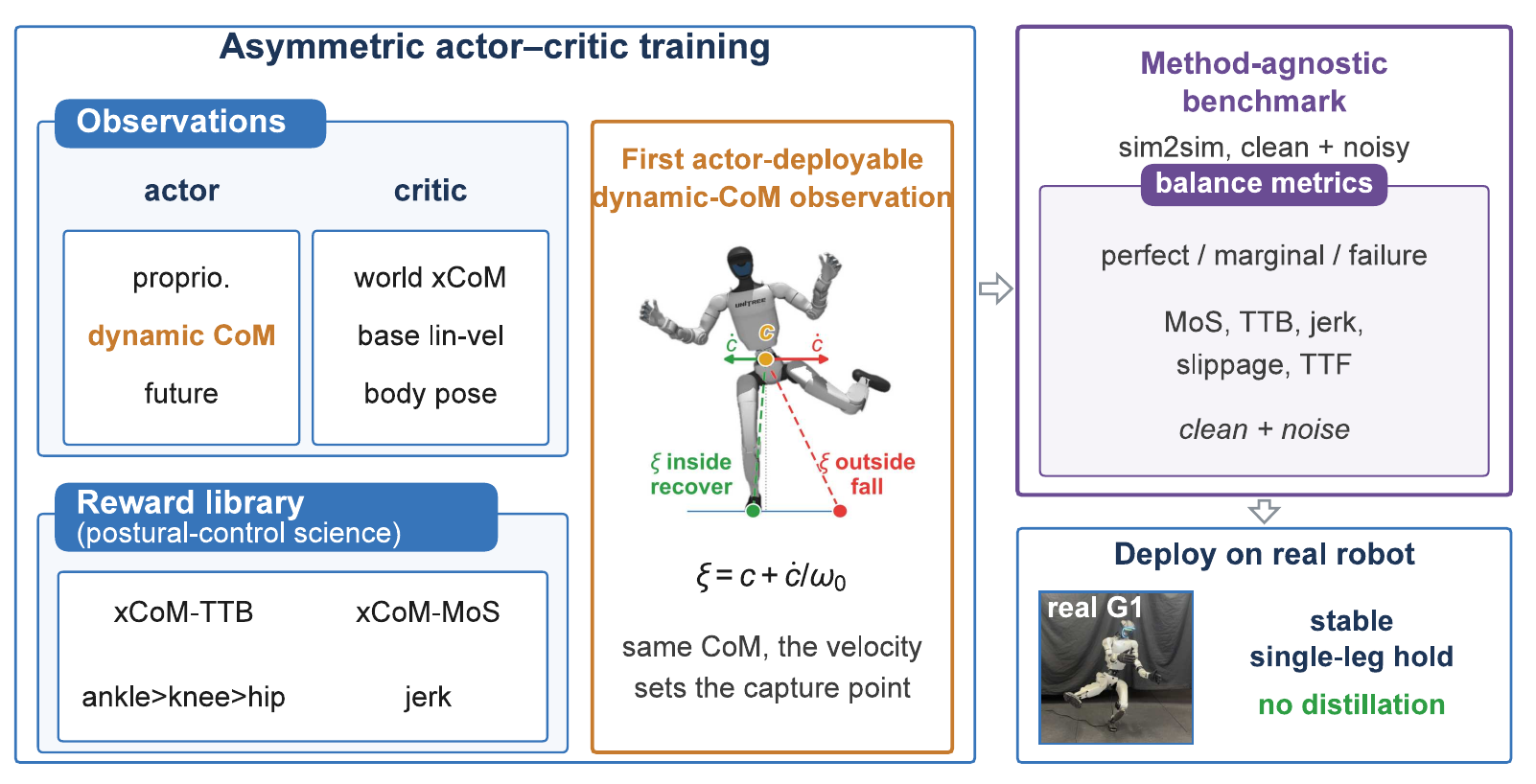}
    \caption{\textbf{Method overview.} One unified single-leg policy is trained with asymmetric FastSAC: a deployable actor (proprioception, a support-relative dynamic-\acs{com} state, and a short future reference) and a privileged critic (state-derived quantities), shaped by a human-science reward library. The same policy is scored on a method-agnostic MuJoCo benchmark and deployed directly to a real Unitree G1.}
    \label{fig:pipeline}
\end{figure}

\Cref{fig:pipeline} overviews the method: an actor that observes the capture point on-board (\cref{sec:obs}), a reward library drawn from postural control (\cref{sec:reward}), and asymmetric training without distillation (\cref{sec:training}). The benchmark that scores the result follows in \cref{sec:benchmark}.

\subsection{A Support-Relative Capture Point}\label{sec:obs}

Single-leg balance is, to first order, a \ac{lip} problem~\citep{kajita2001,pratt2006}. Modeling the whole-body \ac{com} $\CoM$ over the stance foot as a pendulum of height $h$, the quantity that decides whether the robot can keep its balance is the capture point, or \ac{xcom},
\begin{equation}
    \xcom = \CoM + \dCoM/\wnat, \qquad \wnat = \sqrt{g/h},
    \label{eq:xcom}
\end{equation}
and the \ac{com} comes to rest over the foot if and only if $\xcom$ lies inside the support polygon~\citep{hof2005,pratt2006}. Since the foot is nearly stationary in single stance, taking \cref{eq:xcom} relative to the support-foot center $\supc$ gives $\xcom - \supc \approx \rrel + \drrel/\wnat$ with $\rrel \triangleq \CoM - \supc$, so the pair $(\rrel, \drrel)$ is the support-relative capture-point state. Position alone does not suffice: a \ac{com} at the edge of the foot is safe if it is moving inward and lost if it is moving outward, and on the Unitree G1 the lateral support half-width is under 3\,cm. It is the velocity $\drrel$ that separates a recoverable state from an imminent fall, and a single-frame policy cannot recover it from position.

\paragraph{The base velocity cancels.}
At first sight $\drrel$ is unmeasurable on hardware, because the \ac{com} velocity contains the base linear velocity, which no on-board sensor provides. The relative formulation removes exactly this term. Let $\bm{R}\in\mathrm{SO}(3)$ and $\bm{p}_b$ be the base (pelvis) orientation and position, $\bm{v}_b = \dot{\bm{p}}_b$ the base linear velocity, and $\bm{\omega}$ the base angular velocity, whose base-frame value $\bm{\omega}^{B} = \bm{R}^{\top}\bm{\omega}^{W}$ is the gyroscope reading. Link $i$ has mass $m_i$, with $M = \sum_i m_i$, and a base-frame position $\bm{d}_i(\bm{q})$ from forward kinematics, so its world position is $\bm{p}_i^{W} = \bm{p}_b^{W} + \bm{R}\,\bm{d}_i(\bm{q})$. The \ac{com} and the support center in the base frame are
\begin{equation}
    \bm{d}_c = \tfrac{1}{M}\textstyle\sum_i m_i\,\bm{d}_i, \qquad
    \bm{d}_s = \textstyle\sum_{i\in\mathcal{C}} w_i\,\bm{d}_{\mathrm{foot},i}, \quad \textstyle\sum_{i\in\mathcal{C}} w_i = 1,
    \label{eq:dcds}
\end{equation}
with $\mathcal{C}$ the set of feet in contact. Differentiating $\bm{p}_i^{W}$ with $\dot{\bm{R}}\bm{d}_i = \bm{\omega}^{W}\times(\bm{R}\bm{d}_i)$ and $\dot{\bm{d}}_i = \bm{J}_i(\bm{q})\dot{\bm{q}}$ gives $\dot{\bm{p}}_i^{W} = \bm{v}_b^{W} + \bm{\omega}^{W}\times(\bm{R}\bm{d}_i) + \bm{R}\bm{J}_i\dot{\bm{q}}$. Mass-averaging over links, doing the same for the support center with $\Jc = \tfrac{1}{M}\sum_i m_i\bm{J}_i$ and $\Js = \sum_{i\in\mathcal{C}} w_i\bm{J}_{\mathrm{foot},i}$, and subtracting, the base velocity, common to every body, drops out:
\begin{equation}
    \drrel^{W} = \dCoM^{W} - \dot{\supc}^{W}
    = \cancel{\bm{v}_b^{W}} - \cancel{\bm{v}_b^{W}}
    + \bm{\omega}^{W}\times\big(\bm{R}(\bm{d}_c - \bm{d}_s)\big) + \bm{R}(\Jc - \Js)\dot{\bm{q}}.
    \label{eq:cancel}
\end{equation}
Rotating into the base frame gives the pair the actor observes,
\begin{equation}
    \boxed{\;\rrel^{B} = \bm{d}_c - \bm{d}_s, \qquad
    \drrel^{B} = \bm{\omega}^{B}\times(\bm{d}_c - \bm{d}_s) + (\Jc - \Js)\,\dot{\bm{q}}\;}
    \label{eq:obs}
\end{equation}
Every term depends only on the encoders $(\bm{q},\dot{\bm{q}})$, which give $\bm{d}_c,\bm{d}_s,\Jc,\Js$ through forward kinematics and the known link masses, and on the gyroscope; neither $\bm{v}_b$ nor $\bm{p}_b$ appears, and the \ac{imu} gravity direction fixes the horizontal projection and the choice of support foot exactly as in training. The cancellation holds for any mass-weighted forward-kinematic point. We take $\bm{d}_c$ over link origins, a proxy $\pCoM$ for the true \ac{com} that omits each link's intra-link offset and is computed by the same rule in simulation and on the robot; over the deployed policy's test rollouts it sits $2.3$\,cm from the true \ac{com} on average, almost entirely fore--aft, with a signed \ac{ml} bias below $1$\,mm (\cref{tab:proxy-err}). In simulation we read per-body world velocities from the engine instead; the two computations are algebraically equivalent under the same nominal model and a fixed support mask, so a policy trained on one consumes the other unchanged, and the residual sim-to-real gap is the ordinary proprioceptive kind, absorbed by observation noise during training (\cref{app:cancel}).

\paragraph{Actor and critic.}
The actor observes $o_{\mathrm{bal}} = (\rrel^{B}, \drrel^{B}) \in \mathbb{R}^{4}$, the horizontal components of \cref{eq:obs}, together with standard proprioception (joint positions and velocities, base angular velocity, projected gravity, and the previous action), the motion command, and a five-frame future-reference window: 463 dimensions, all available on-board (\cref{tab:actor-obs}). The future window mirrors anticipatory postural adjustment~\citep{massion1992}, letting the single-frame policy prepare for the swing foot's lift or a support switch rather than correct after it. A privileged critic shares the actor's command, proprioceptive, and future-reference inputs, but replaces $o_{\mathrm{bal}}$ with the true whole-body xCoM relative to the support center and expressed in the base frame. It additionally observes the base linear velocity, rigid-body positions and orientations, and the motion-reference position, all expressed in the base frame. The critic is used only for value learning during training and is discarded at deployment (Sec. B).

\subsection{A Human-Science Reward Library}\label{sec:reward}

\begin{table}[t]
    \caption{\textbf{The human-science reward library.} Each term reads or steers the same support-relative capture point the actor observes (\cref{sec:obs}); weights are listed in \cref{tab:hyperparams}.}
    \label{tab:reward}
    \centering
    \small
    \setlength{\tabcolsep}{3pt}
    \begin{tabular}{@{}>{\raggedright\arraybackslash}p{0.19\linewidth}>{\raggedright\arraybackslash}p{0.47\linewidth}>{\raggedright\arraybackslash}p{0.3\linewidth}@{}}
        \toprule
        Reward term & Role & Human-science basis \\
        \midrule
        Capture-point margin (\acs{mos}) & soft spatial buffer: penalize the \acs{xcom} within a safety margin (2\,cm single / 3\,cm double support) of the support boundary & \acs{mos}~\citep{hof2005} \\[2pt]
        \acs{ttb} & soft temporal buffer: penalize once the \acs{xcom}'s projected time to cross the boundary drops below a reaction threshold (0.3\,s single / 0.2\,s double support) & \acs{ttb}~\citep{hertel2006} \\[2pt]
        Ankle$\to$knee$\to$hip rate & graded stance-leg action-rate cost (ankle penalized most, knee less, hip free) so the ankle holds the \acs{com} steady rather than chasing it & distal-first (ankle) postural strategy~\citep{tropp1988,riemann2003} \\[2pt]
        Jerk penalty & penalize action jerk, suppressing high-frequency chatter that is nominally stable but undeployable & jerk indexes postural-control quality~\citep{semak2020} \\
        \bottomrule
    \end{tabular}
\end{table}

Human postural control supplies principles that humanoid balance rewards have largely not used. We translate four of them into penalty terms (\cref{tab:reward}), each read on the same support-relative capture point the actor observes. Two are soft margins on the capture point. A spatial one, the \ac{mos}~\citep{hof2005}, penalizes the \ac{xcom} once it comes within a small margin of the single-foot support boundary; a temporal one, the \ac{ttb}~\citep{hertel2006}, penalizes it once its projected time to cross that boundary falls below a reaction threshold, ``how much time is left to react.'' Both are soft rather than hard constraints: a hard boundary would fight the dynamic reference, whereas a soft margin preserves a recoverable buffer without forbidding motion. The third term encodes the ankle$\to$knee$\to$hip hierarchy by which humans reject small perturbations at the ankle and recruit the knee and hip only as the disturbance grows~\citep{tropp1988,riemann2003}: graded action-rate penalties on the stance leg, heaviest on the ankle, lighter on the knee, and none on the hip beyond the global action-rate term. Penalizing the rate rather than the use of the ankle steers it toward smooth, sustained torque that holds the \ac{com} in place instead of chasing it. The fourth penalizes action jerk, the second difference of the action, following evidence that jerk indexes postural-control quality~\citep{semak2020}; it suppresses the high-frequency chatter by which a policy could hold balance nominally while being undeployable. Standard tracking, contact, and regularization terms complete the reward.

\subsection{Training Without Distillation}\label{sec:training}

We train a single policy on the stratified motion set of \cref{sec:dataset} with FastSAC~\citep{seo2025}, an asymmetric distributional soft actor--critic, on the Holosoma framework~\citep{holosoma2025}. The actor sees only the on-board observation above; the critic additionally sees the privileged block. There is no teacher and no student: the actor that trains is the actor that deploys, exported unchanged to the robot. Domain randomization covers friction, link masses, the base-\ac{com} offset, random pushes, and additive observation noise, including on the capture-point channel; training runs 8192 parallel environments in IsaacSim at 50\,Hz control (\cref{app:training}).

\section{A Method-Agnostic Balance Benchmark}\label{sec:benchmark}

Progress on single-leg balance is hard to measure. Specialized methods lack a shared instrument, and the general motion-tracking policies increasingly used as humanoid skill substrates are never tested for single-leg competence at all. We build a MuJoCo benchmark that scores any released policy from its physical outcome alone, with a metric suite grounded in postural control. It serves three uses of widening scope: selecting the checkpoint to deploy from a single training run (\cref{sec:ckpt}), comparing methods under one protocol (\cref{sec:results}), and probing the single-leg competence of any generalist, zero-shot or after co-training on our motion set.

\subsection{A Stratified Motion Set}\label{sec:dataset}

Our motions come from the AMS synthetic balance-motion dataset~\citep{ams2026}. From its 9,814 valid clips we draw a compact, class-balanced random subsample of 900 (seed-fixed); each is a fixed-length 4.98\,s trajectory that passes from double to single and back to double support, and its middle single-support window (${\approx}3.3$\,s) is the phase the benchmark scores. We stratify by two orthogonal pose axes, support-leg squat depth (pelvis height) and swing-foot absolute height, into a $3\times3$ grid of nine classes with 100 clips each, balanced between left and right support. The set is split 720 train / 90 validation / 90 test, balanced across classes and sides; checkpoints are selected on validation and every number is reported on test. Difficulty is not asserted a priori but emerges on the grid: the deep-squat, high-lift corner is hardest and the tall-pelvis, low-lift corner easiest for prior policies (\cref{sec:results}). Axis choice, bin edges, and the leak-free split are detailed in \cref{app:dataset}.

\subsection{Protocol and Metrics}\label{sec:protocol}

\paragraph{Method-agnostic at the command interface.}
Every policy is scored on the same motions by the same metric, read from the robot's physical state rather than from any method's internal reference or reward. A thin adapter builds each method's documented observation from the state and the reference, runs its network, and decodes its actions to joint targets at the robot's low-level state-to-command contract; methods whose exact specification is unavailable are excluded rather than guessed. Evaluation uses MuJoCo-based stacks. Physics harnesses differ across methods: four baselines reuse our G1 plant and four run in their own released deployment harness, each cleared by a tracking sanity check first (\cref{app:deploy}). Tier outcomes therefore rest on the shared true-state metric rather than on an identical physics plant, and continuous values across methods are descriptive rather than a strict ranking.

\paragraph{Outcome tiers.}
Each trial falls into one of three mutually exclusive tiers. A \emph{Perfect} hold is a clean single-leg stance across the single-support window: the support foot never hops, the swing foot never touches down, the robot does not fall, and the body tracks the reference within HuB's average 12-keypoint 0.5\,m gate~\citep{hub2025}. A \emph{Marginal} success does not fall but stays upright only by breaking that constraint, hopping the support foot or touching the swing foot down; both are the same move, a base-of-support reconfiguration that catches an escaped capture point, and neither is balance. A \emph{Failure} is a fall.

\paragraph{Continuous diagnostics and conditions.}
Alongside the tiers we report, on the balance axis, the \ac{mos} of the true whole-body \ac{com} and the time the capture point spends outside the support polygon, plus HuB's keypoint tracking errors, support-foot slippage, a jerk proxy, and \ac{ttf}, all computed from each policy's resulting state (\cref{app:metrics-def}). We report two conditions: a clean, deterministic run ($K{=}1$, no observation noise), used for cross-method comparison with each baseline in its most favorable setting, and a noisy run ($K{=}10$ seeds) with deployment-relevant observation noise, a temporally correlated Ornstein--Uhlenbeck \ac{imu}-orientation error on the gyroscope and projected gravity plus per-step joint-velocity noise with a one-step delay, used for robustness and checkpoint selection. Noise perturbs observations only; tiers are always judged on the true state. We adopt HuB's metric definitions for comparability but do not place our numbers beside HuB's, since the simulators and task sets differ.

\section{Experiments}\label{sec:exp}

Every baseline is a released, general-purpose humanoid whole-body motion-tracking policy; none was built for balance. We include every such policy we could obtain and run on the G1 that passes a multi-clip sanity check, eight entries spanning 2024--2026: ProtoMotions~\citep{protomotions2024}, GMT~\citep{gmt2025}, TWIST~\citep{twist2025}, SONIC~\citep{sonic2025}, OmniXtreme~\citep{omnixtreme2026}, MOSAIC~\citep{mosaic2026}, Humanoid-GPT~\citep{humanoidgpt2026}, and HoloMotion~\citep{holomotion2026}. These are strong generalists, not weak baselines; each is scored at its native command interface and action dimensionality (29~\acp{dof}, or 23 for GMT and TWIST), clean and noise-free. \method is trained as in \cref{sec:training}, with all hyperparameters in \cref{app:training}.

\subsection{Results}\label{sec:results}

\begin{table}[t]
    \caption{\textbf{Single-leg balance results.} Held-out test set, clean and deterministic, $n{=}90$ clips per method; \method is the deployed checkpoint (\cref{sec:ckpt}). Outcome tiers are mutually exclusive (\cref{sec:protocol}). \acs{mos}: capture-point margin on the true whole-body \acs{com} (\acs{ml}\,/\,\acs{ap}; less negative is better; per-motion minimum over the single-support window, averaged over the 90 clips). \acs{xcom}-out: time the capture point spends outside the support polygon (window ${\approx}3.3$\,s). The full continuous suite is in \cref{tab:metrics-full}. OmniXtreme is stochastic; a fixed default-seed sample is shown, with its Perfect tier seed-invariant at 0/90 (\cref{app:deploy}). Best in \textbf{bold}.}
    \label{tab:main}
    \centering
    \small
    \begin{tabular}{@{}l ccc cc@{}}
        \toprule
        & \multicolumn{3}{c}{Outcome (\%)} & \multicolumn{2}{c}{Balance} \\
        \cmidrule(lr){2-4}\cmidrule(lr){5-6}
        Method & Perfect\,$\uparrow$ & Marginal & Failure\,$\downarrow$ & \acs{mos}$_{\text{\acs{ml}/\acs{ap}}}$\,$\uparrow$ & \acs{xcom}-out\,$\downarrow$ \\
        \midrule
        \textbf{\method (ours)} & \textbf{98.9} & 0.0 & \textbf{1.1} & \textbf{$-$0.025\,/\,$+$0.048} & \textbf{0.09} \\
        \midrule
        SONIC & 0.0 & 81.1 & 18.9 & $-$0.281\,/\,$-$0.209 & 2.92 \\
        Humanoid-GPT & 0.0 & 77.8 & 22.2 & $-$0.218\,/\,$-$0.092 & 3.08 \\
        HoloMotion & 0.0 & 75.6 & 24.4 & $-$0.304\,/\,$-$0.174 & 3.07 \\
        MOSAIC & 0.0 & 66.7 & 33.3 & $-$0.376\,/\,$-$0.270 & 2.98 \\
        ProtoMotions & 0.0 & 51.1 & 48.9 & $-$0.314\,/\,$-$0.250 & 3.06 \\
        TWIST & 0.0 & 47.8 & 52.2 & $-$0.615\,/\,$-$0.527 & 3.24 \\
        GMT & 0.0 & 18.9 & 81.1 & $-$0.660\,/\,$-$0.460 & 3.19 \\
        OmniXtreme & 0.0 & 5.6 & 94.4 & $-$0.736\,/\,$-$0.611 & 3.25 \\
        \bottomrule
    \end{tabular}
\end{table}

\begin{wrapfigure}{r}{0.5\linewidth}
    \centering
    \includegraphics[width=\linewidth]{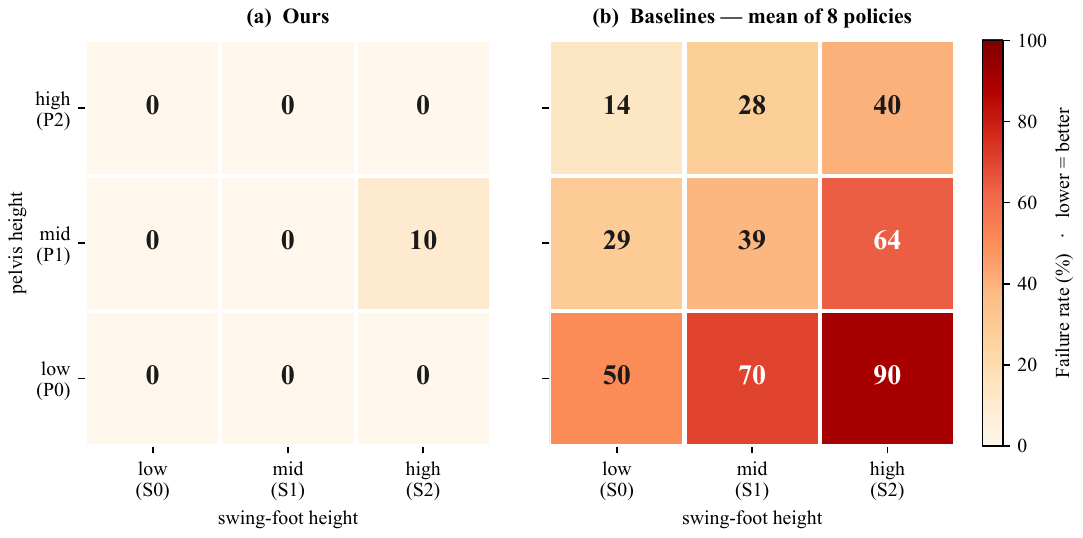}
    \caption{\textbf{Per-class failure rate.} Fall rate over the $3\times3$ pose grid (ten held-out clips per class; clean). (a) \method falls in one clip of one class. (b) The eight-baseline mean rises from the easy corner (high pelvis, low swing foot) to the hard one (deep squat, high swing foot).}
    \label{fig:heatmap}
\end{wrapfigure}

\paragraph{\method holds single-leg balance.}
\method holds a clean single-leg stance on 89 of 90 held-out motions (98.9\% Perfect, 0\% Marginal, 1.1\% Failure; \cref{tab:main}), and the continuous suite shows that the hold is biomechanically genuine on every axis (\cref{tab:metrics-full}). Measured on the true \ac{com}, its capture point rides the edge of the support polygon, with a positive \ac{ap} margin (fore--aft) and a near-zero \ac{ml} one (\ac{mos} $-0.025$\,/\,$+0.048$), and crosses the boundary for only 0.09\,s of the ${\approx}3.3$\,s window. Contact and tracking are clean, control is smooth, and the one miss is a single clip in the class that pairs a high swing foot with a mid-height pelvis (\cref{fig:heatmap}a).

\paragraph{Generalists recover; they do not balance.}
Under the same clean protocol, all eight released generalists reach a Perfect rate of exactly 0/90. The zero is not because they fall: most rarely fall and instead land in Marginal, staying upright by re-planting the swing foot and hopping the support foot, and 88.5\% of Marginal clips show both moves at once. Two profiles emerge. Fall-dominant policies fail outright (OmniXtreme 94.4\% and GMT 81.1\% Failure, Marginal ${\le}18.9\%$), whereas recovery-dominant ones rarely fall but survive by this coupled shuffle (Humanoid-GPT 22.2\% and HoloMotion 24.4\% Failure, Marginal ${\geq}75.6\%$). The strongest is SONIC, a large tracking foundation model with no capture-point or support-polygon design~\citep{sonic2025}: it falls least (18.9\%), stays up longest (\ac{ttf} 4.6\,s), and reaches Marginal on 81\% of trials, so a sufficiently strong generalist begins to show emergent single-leg robustness. The emergence is partial. SONIC never achieves a single Perfect hold, and its capture-point margin is more than $10\times$ worse than \method's (\ac{mos} $-0.28$ \vs $-0.025$): scale confers robustness, not balance.

The continuous metrics tell the same story at substantial gaps (\cref{tab:metrics-full}): the generalists' capture-point margin is deeply negative on both axes (\ac{mos} $-0.74$ to $-0.09$) against \method's near-zero and only positive \ac{ap} margin; their support foot slips at 54--378\,mm/s against 4\,mm/s; their tracking-fail rate is 16--96\%; and their control is 3--11$\times$ jerkier. Difficulty emerges along both pose axes (\cref{fig:heatmap}b): the eight-baseline failure rate rises smoothly from ${\sim}15\%$ (tall pelvis, low swing foot) to ${\sim}90\%$ (deep squat, high swing foot), where a high swing foot forces the \ac{com} over the support foot at its least stable. \method achieves 9/10 Perfect holds in its only sub-perfect class, with one failed rollout. Where the baselines collapse most, \method still holds.

\subsection{Ablations}\label{sec:ablation}

\begin{wraptable}{r}{0.5\linewidth}
    \vspace{-36pt}
    \caption{\textbf{Ablations.} Perfect-success rate (\%) on the held-out test set as each component is removed from the full \method policy (or, for the observation, degraded) and the checkpoint re-selected under the same protocol, clean ($K{=}1$) and under deployment-relevant noise ($K{=}10$, ten seeds; \cref{sec:protocol}). The dynamic-\acs{com} observation dominates; the smaller-margin components are directionally positive in these single-run ablations. Best in \textbf{bold}.}
    \label{tab:ablation}
    \centering
    \small
    \setlength{\tabcolsep}{4pt}
    \begin{tabular}{@{}l cc@{}}
        \toprule
        Ablation & Clean & Noisy \\
        \midrule
        \textbf{\method (ours)} & \textbf{98.9} & \textbf{61.8} \\
        \midrule
        w/o future-reference obs. & 93.3 & 57.3 \\
        w/o ankle action-rate penalty & 86.7 & 47.7 \\
        w/o knee action-rate penalty & 85.6 & 44.0 \\
        w/o \acs{mos} penalty & 83.3 & 51.0 \\
        w/o jerk penalty & 82.2 & 48.1 \\
        w/o \acs{ttb} penalty & 73.3 & 43.8 \\
        dynamic $\to$ static \acs{com} obs. & 66.7 & 29.1 \\
        w/o \acs{com} obs. & 55.6 & 8.6 \\
        \bottomrule
    \end{tabular}
\end{wraptable}

We remove each component from the full \method policy (or, for the observation, degrade it), re-select the checkpoint under the same protocol, and report Perfect success clean and under the deployment-relevant noise (\cref{tab:ablation}). Every component is net-positive, so none is redundant, but their contributions are sharply unequal.

\paragraph{The dynamic-\acs{com} observation is the largest driver.}

Removing it drops Perfect success from 98.9\% to 55.6\% ($-43$ points), and merely making it static, position without velocity, already costs 32 points (to 66.7\%). This is the direct test of the paper's claim: it is not enough to reward the capture point; the actor must observe it, and observe it dynamically. Under deployment noise the effect is starker still, collapsing to 8.6\% ($-53$ points): \ac{imu} noise corrupts the orientation estimate, and only the dynamic-\ac{com} observation tells the policy which way it is actually toppling. The gap is already visible in training, where the two \ac{com}-deprived runs plateau at 382 and 416 steps of the 500-step episode against 465--490 for every other run (\cref{fig:train-eplen}). Even with the critic seeing the privileged balance state, an actor blind to its own capture point cannot fit these motions, so the observation is a prerequisite for learning the task, not only a deployment-time advantage.

\begin{wrapfigure}[18]{r}{0.4\linewidth}
    \centering
    \includegraphics[width=\linewidth]{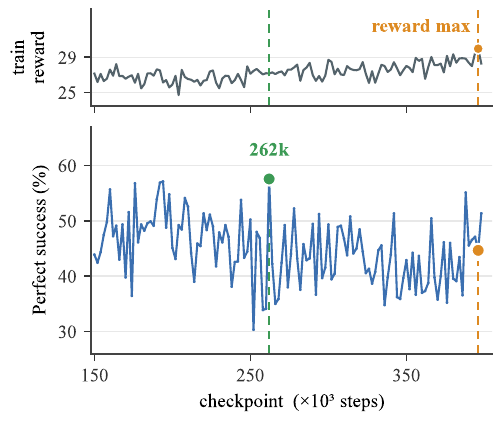}
    \caption{\textbf{Reward does not rank checkpoints.} Training reward (top) saturates while validation Perfect success (bottom) swings between neighboring checkpoints; the deployed checkpoint (green) is far from both the last and the highest-reward one (amber).}
    \label{fig:ckpt-select}
\end{wrapfigure}

\paragraph{The margin rewards come next.}
The soft \ac{ttb} reward is the second driver ($-26$ points, to 73.3\%), and its spatial counterpart, the \ac{mos} penalty, costs a further 16 points: the two complementary capture-point margins guard balance in time and in space. The jerk penalty costs 17 points, so smoothness is material to the hold itself and not only a deployability safeguard; the knee and ankle action-rate penalties cost 13 and 12 points, the two hierarchy terms weighing in nearly equally. The future-reference window contributes least ($-5.6$ clean, $-4.5$ noisy), consistent with its anticipatory role.

\subsection{Reward Does Not Rank Checkpoints}\label{sec:ckpt}

Every \method number above comes from one checkpoint, chosen by the benchmark rather than by training reward. A training run emits hundreds of checkpoints, and their training reward saturates once the tracking objective fits and is essentially uncorrelated with sim2sim Perfect success (\cref{fig:ckpt-select}, top), while validation Perfect success swings widely between neighboring checkpoints (bottom). Neither the last checkpoint nor the highest-reward one is therefore a safe choice: the measured-best checkpoint, at step 262k, is neither the final nor the highest-reward one. Sweeping every checkpoint on the benchmark under the noisy protocol, and taking the one that measurably balances best, is the checkpoint-selection use of the benchmark. We deploy this checkpoint and report it throughout.

\subsection{Real-Robot Deployment}\label{sec:deploy}

\begin{figure}[t]
    \centering
    \includegraphics[width=\linewidth]{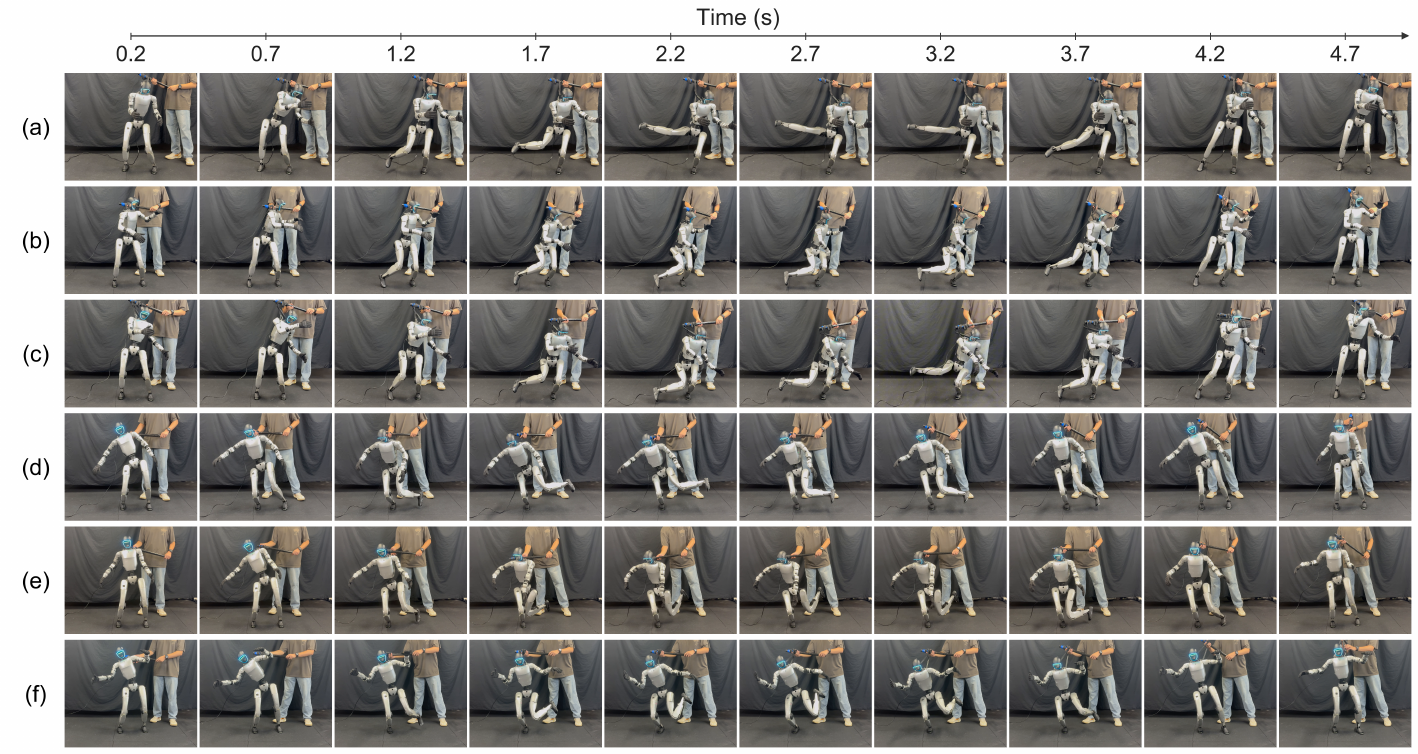}
    \caption{\textbf{Real-robot deployment over time.} Rows (a)--(f) show six representative real-robot rollouts of the same unified \method policy, each tracking a distinct single-leg reference motion. Columns show ten snapshots sampled every 0.5\,s from $t=0.2$ to $4.7$\,s, spanning the transition from double to single and back to double support. The visible operator and safety tether are used only for fall protection and provide no mechanical assistance during the displayed trials.}
    \label{fig:realrobot}
\end{figure}

We deploy the benchmark-selected checkpoint on a physical Unitree G1 (29 \acp{dof}), running the same actor directly (ONNX, 50\,Hz) with every observation reconstructed on-board and no privileged state. The dynamic-\ac{com} observation is computed from the encoder joint positions and velocities, the \ac{imu} gyroscope, and projected gravity alone; the base's absolute position and velocity are held at zero in the state buffer, where the support-relative differencing cancels them exactly, so the actor receives the same quantity as in simulation, and the support foot is chosen by the same gravity-aligned foot-height rule as in training, so the support switch has no train-to-deploy discontinuity. Qualitative hardware trials show that the robot maintains single-leg balance across diverse reference motions. \Cref{fig:realrobot} presents six representative examples, with additional demonstration videos available on the \href{https://estoil.github.io/DDC/}{project page}, demonstrating that the design transfers to hardware. Deployment details are in \cref{app:deploy}.

\section{Limitations and Conclusion}\label{sec:conclusion}

\paragraph{Limitations.}
Our quantitative evaluation is sim2sim; the real-robot results confirm deployability only qualitatively. Real-world metrics, ports of the benchmark beyond the Unitree G1, and giving this root-cause balance competence to general policies, by co-training or fine-tuning on our motion set, are future work.

\paragraph{Conclusion.}
Unified humanoid policies, for all their agility, do not hold single-leg balance: they recover by stepping rather than prevent imbalance. Making the capture point proprioceptive through a change of frame, and pairing the resulting support-relative dynamic-\ac{com} observation with a human-science reward library, yields a policy that holds a clean stance across a stratified pose grid and transfers to a real Unitree G1 without distillation. We release the data, code, policy, and benchmark so that single-leg balance becomes a capability the field can measure and build in rather than a per-task demonstration.

\subsection*{Reproducibility Statement}

The observation derivation is given in \cref{sec:obs}, with its remaining details in \cref{app:cancel} and the full actor and critic inputs in \cref{app:obs}. Training configuration, dataset construction and the leak-free split, metric definitions and thresholds, and the evaluation and deployment protocols are documented in \cref{app:training,app:dataset,app:metrics-def,app:deploy}. The code, deployed policy checkpoint, motion dataset, and benchmark harness are publicly available through the \href{https://estoil.github.io/DDC/}{project page}.

\section*{Acknowledgments}

This work was supported by the Fundamental Research Funds
for the Central Universities, Peking University.

\bibliography{reference_header,references}
\bibliographystyle{iclr2027_conference}

\clearpage
\appendix

\renewcommand\thefigure{A\arabic{figure}}
\setcounter{figure}{0}
\renewcommand\thetable{A\arabic{table}}
\setcounter{table}{0}
\renewcommand\theequation{A\arabic{equation}}
\setcounter{equation}{0}
\setcounter{footnote}{0}

\section{Support-Relative Capture Point: Remarks and Reconstruction}\label{app:cancel}

\paragraph{Frame of the velocity.}
$\drrel^{B}$ in \cref{eq:obs} denotes the world-frame relative velocity $\dCoM^{W}-\dot{\supc}^{W}$ expressed in base coordinates, $\bm{R}^{\top}(\dCoM^{W}-\dot{\supc}^{W})$, not the coordinate time derivative of $\rrel^{B}$; the $\bm{\omega}^{B}\times$ term is exactly the rotating-frame correction between the two.

\paragraph{Support switches.}
The support center holds the contact weights $w_i$ fixed within a support phase (in single support, $w{=}1$), so $\dot{w}_i=0$ and no $\dot{w}_i\,\bm{d}_{\mathrm{foot},i}$ term appears in \cref{eq:cancel}. At a support-mask switch this no longer holds and the observed pair can step. The actor sees the observation through these switches, not only inside the single-support window used for evaluation, and the 3\,cm double-support band, which blends both feet in $\bm{d}_s$, smooths the switch rather than removing it.

\paragraph{On-board reconstruction.}
Every term in \cref{eq:obs} depends only on the joint encoders $(\bm{q},\dot{\bm{q}})$, which give $\bm{d}_c$, $\bm{d}_s$, $\Jc$, and $\Js$ through forward kinematics and the known link masses, and on the gyroscope $\bm{\omega}^{B}$; the \ac{imu} gravity direction fixes the horizontal projection and the gravity-aligned choice of support foot. Neither the base linear velocity $\bm{v}_b$ nor the absolute base position $\bm{p}_b$ appears. In simulation we instead read the engine's per-body world velocities; the two computations are algebraically equivalent under the same nominal kinematic and mass model and a fixed support mask, so a policy trained on the simulator value consumes the on-robot reconstruction unchanged. The residual sim-to-real gap is the ordinary proprioceptive kind, encoder-velocity noise, kinematic and mass-model error, and contact-set transients, which the domain-randomization noise on this channel absorbs during training. \Cref{tab:proxy-err} quantifies the gap between the link-origin proxy the actor observes and the true \ac{com} on which every balance metric is computed.

\begin{table}[b]
    \caption{\textbf{Deployable proxy \boldmath$\pCoM$ versus the true whole-body \acs{com} \boldmath$\CoM$.} Statistics over the deployed policy's test-90 rollouts (14,820 single-support frames). The offset is at the centimeter scale, comparable to the ${\approx}2.75$\,cm lateral support half-width, so all balance metrics (\cref{tab:main,tab:metrics-full}) are computed on the true \acs{com}. The bias is mostly fore--aft; on the \acs{ml} axis, the binding one for single-leg stance, the signed bias is near zero. The \acs{xcom} error holds the \acs{com} height fixed to the true value, isolating the horizontal position and velocity difference. The 179.4 mm maximum occurs in the sole failed rollout, during a brief flight/fall transition within the reference right-support window. It is dominated by the velocity-extrapolation term (146.5 mm), while the instantaneous horizontal position offset is 33.1 mm. Excluding this failed rollout, the maximum over the remaining 89 motions is 56.7 mm.}
    \label{tab:proxy-err}
    \centering
    \small
    \begin{tabular}{@{}l ccc@{}}
        \toprule
        Quantity (mm) & Mean & P95 & Max \\
        \midrule
        3D \acs{com} position error & 22.5 & 36.6 & 53.2 \\
        Horizontal \acs{xcom} error & 20.8 & 35.2 & 179.4 \\
        \midrule
        \acs{ap} signed bias & \multicolumn{3}{c}{$+17.8$} \\
        \acs{ml} signed bias & \multicolumn{3}{c}{$+0.94$ ($|\cdot|$ mean 6.0, P95 13.9)} \\
        \bottomrule
    \end{tabular}
\end{table}

\section{Observation Design}\label{app:obs}

The deployed actor observation $o_t=(o_{\mathrm{bal}},\,o_{\mathrm{prop}},\,o_{\mathrm{cmd}},\,o_{\mathrm{fut}})$ is a 463-dimensional vector, every term of which is available on-board without privileged state: the balance and proprioceptive terms come from the joint encoders and the \ac{imu} (pelvis frame), the command and future-reference terms from the loaded motion, and the previous action from the controller (\cref{tab:actor-obs}). Its key term is the support-relative dynamic-\ac{com} $o_{\mathrm{bal}}$, computed on the deployable link-origin proxy (\cref{sec:obs,app:cancel}). The privileged critic retains the actor's command, proprioceptive, and future-reference inputs, but replaces the actor's four-dimensional, proxy-based $o_{\mathrm{bal}}=(r^B,\dot{r}^B)$ with a two-dimensional true whole-body xCoM relative to the support center and expressed in the base frame. It shapes value during training only and is discarded at deployment, so the actor deploys directly.

\paragraph{Why a future window.}
$o_{\mathrm{fut}}$ mirrors anticipatory postural adjustment~\citep{massion1992}: before a self-generated movement, such as the swing foot's lift or a switch between single and double support, the nervous system pre-activates postural muscles to compensate for the coming disturbance rather than correct it afterward. Exposing the upcoming reference lets the otherwise single-frame policy prepare instead of chasing the error; its ablation in \cref{sec:ablation} confirms that it matters.

\begin{table}[t]
    \caption{\textbf{Actor and critic observations.} Left: the deployed actor observation $o_t$, one row per term, with bracketed per-block totals. Right: the privileged critic observation; the state-derived block is training-only and discarded at deployment.}
    \label{tab:actor-obs}
    \centering
    \small
    \setlength{\tabcolsep}{5pt}
    \begin{minipage}[t]{0.56\linewidth}
        \centering
        \begin{tabular}[t]{@{}l l r@{}}
            \toprule
            Block & Observation term & Dim \\
            \midrule
            $o_{\mathrm{bal}}$ [4] & dynamic-\acs{com} $(\rrel^{B},\drrel^{B})$ & 4 \\
            \midrule
            $o_{\mathrm{prop}}$ [93] & base angular velocity & 3 \\
            & joint positions & 29 \\
            & joint velocities & 29 \\
            & previous action & 29 \\
            & projected gravity & 3 \\
            \midrule
            $o_{\mathrm{cmd}}$ [66] & motion command & 58 \\
            & motion reference orientation & 6 \\
            & reference support phase & 2 \\
            \midrule
            $o_{\mathrm{fut}}$ [300] & future command ($5\times58$) & 290 \\
            & future support phase ($5\times2$) & 10 \\
            \midrule
            \multicolumn{2}{@{}l}{\textbf{Actor total}} & \textbf{463} \\
            \bottomrule
        \end{tabular}
    \end{minipage}%
    \begin{minipage}[t]{0.44\linewidth}
        \centering
        \begin{tabular}[t]{@{}l r@{}}
            \toprule
            Observation term & Dim \\
            \midrule
            \multicolumn{2}{@{}l}{\emph{Shared with the actor (on-board)}} \\
            motion command & 58 \\
            motion reference orientation & 6 \\
            base angular velocity & 3 \\
            joint positions & 29 \\
            joint velocities & 29 \\
            previous action & 29 \\
            projected gravity & 3 \\
            reference support phase & 2 \\
            future support phase ($5\times2$) & 10 \\
            future command ($5\times58$) & 290 \\
            \midrule
            \multicolumn{2}{@{}l}{\emph{Privileged (training only)}} \\
            support-relative \acs{xcom} & 2 \\
            base linear velocity $\bm{v}_b$ & 3 \\
            rigid-body positions ($14\times3$) & 42 \\
            rigid-body orientations ($14\times6$) & 84 \\
            motion reference position & 3 \\
            \midrule
            \textbf{Critic total} & \textbf{593} \\
            \bottomrule
        \end{tabular}
        \label{tab:critic-obs}
    \end{minipage}
\end{table}

\section{Training Details}\label{app:training}

Our implementation builds on the public Holosoma framework~\citep{holosoma2025}. \Cref{tab:hyperparams} lists the full configuration of the deployed \method run, read from its logs: an asymmetric distributional soft actor--critic~\citep{seo2025} with twin critics, an auto-tuned entropy temperature, observation normalization, and \ac{mlp} actor and critic networks, trained for $4\times10^{5}$ iterations over 8192 parallel IsaacSim environments on two RTX 3090 GPUs. Domain randomization covers friction, restitution, per-link mass, the base-\ac{com} offset, joint startup bias, and random pushes, and additive uniform observation noise is applied to every actor channel, including the capture point.

\begin{figure}[t]
    \centering
    \includegraphics[width=0.5\linewidth]{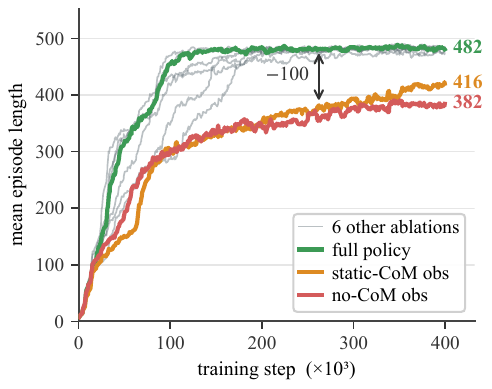}
    \caption{\textbf{Training-time mean episode length.} The full policy and the six other ablations converge near the full 500-step episode; only the two \acs{com}-deprived runs (no \acs{com} observation, static-\acs{com} observation) plateau far short, so without the dynamic-\acs{com} observation the policy never fits the motions.}
    \label{fig:train-eplen}
\end{figure}

\Cref{fig:train-eplen} shows the training-time mean episode length. The full policy and six of the eight ablations converge near the full 500-step episode, whereas the two \ac{com}-deprived runs plateau at 382 and 416 steps: without a dynamic capture-point observation the actor never fits the motions, so the observation is a prerequisite for learning the task and not only for deploying it (\cref{sec:ablation}).

\begin{table}[t]
    \caption{\textbf{Training hyperparameters.} Full configuration of the deployed \method run, trained with FastSAC~\citep{seo2025}.}
    \label{tab:hyperparams}
    \centering
    \small
    \setlength{\tabcolsep}{6pt}
    \begin{tabular}{@{}l l@{}}
        \toprule
        Setting & Value \\
        \midrule
        \multicolumn{2}{@{}l}{\emph{Algorithm: asymmetric distributional soft actor--critic}} \\
        Learning rate (actor / critic / temperature) & $3\times10^{-4}$ \\
        Discount $\gamma$ & 0.99 \\
        Target smoothing $\tau$ & 0.05 \\
        Replay batch size & 8192 \\
        Replay buffer size (per env) & 384 \\
        Gradient updates per env step & 4 \\
        Twin critics ($Q$ networks) & 2 \\
        Distributional atoms / value support & 501 / $[-20, 20]$ \\
        Entropy temperature & auto-tuned (target-entropy ratio 0.5) \\
        Obs.\ normalization / layer-norm / tanh squash & yes \\
        Mixed precision & bf16 \\
        Actor / critic \acs{mlp} width & 512 / 768 \\
        Total training iterations & $4\times10^{5}$ \\
        Random seed & 42 (${+}$ GPU rank under distributed training) \\
        \midrule
        \multicolumn{2}{@{}l}{\emph{Simulation and environment (IsaacSim)}} \\
        Parallel environments & 8192 \\
        Control rate (decimation) & 50\,Hz ($\times4$) \\
        Physics rate & 200\,Hz \\
        Episode length & 10\,s (500 steps) \\
        Robot & Unitree G1, 29 \acs{dof} \\
        Training GPUs & 2$\times$ NVIDIA RTX 3090 \\
        \midrule
        \multicolumn{2}{@{}l}{\emph{Domain randomization}} \\
        Static / dynamic friction & $[0.3, 1.6]$ / $[0.3, 1.2]$ \\
        Restitution & $[0, 0.5]$ \\
        Link-mass scale (per link; base fixed) & $[0.9, 1.1]\times$ \\
        Base-\acs{com} offset ($x$/$y$/$z$) & $[\pm0.025, \pm0.05, \pm0.05]$\,m \\
        Joint-position startup bias & $\pm0.01$\,rad \\
        Random pushes & every 1--3\,s; $|v|\le[0.5,0.5,0.2]$\,m/s, $[0.52,0.52,0.78]$\,rad/s \\
        Action delay (training) & disabled \\
        Actuator \acs{pd} gains / residual force injection & disabled \\
        \midrule
        \multicolumn{2}{@{}l}{\emph{Observation noise (training; additive uniform, $\pm$)}} \\
        Joint velocity & 0.5 \\
        Base angular velocity & 0.2 \\
        Projected gravity & 0.03 \\
        Joint position & 0.01 \\
        Motion-reference orientation & 0.05 \\
        Capture point (\acs{com} relative to support) & 0.015 \\
        \midrule
        \multicolumn{2}{@{}l}{\emph{Reward weights}} \\
        Balance: capture-point margin (\acs{mos}) & $-20$ \\
        Balance: \acs{ttb} & $-15$ \\
        Balance: stance ankle / knee action-rate & $-0.3$ / $-0.1$ \\
        Balance: action jerk (second action difference) & $-0.1$ \\
        Contact: support-contact mismatch & $-2$ \\
        Contact: support-foot slip & $-1$ \\
        Contact: undesired contacts & $-0.1$ \\
        Tracking: body position (rel.\ / global) & $2.0$ / $1.0$ \\
        Tracking: body orientation (rel.\ / global) & $1.0$ / $0.5$ \\
        Tracking: body linear / angular velocity & $1.0$ / $1.0$ \\
        Regularization: action rate ($L_2$) / joint limits & $-1.0$ / $-10.0$ \\
        \bottomrule
    \end{tabular}
\end{table}

\clearpage

\section{Dataset Details}\label{app:dataset}

\paragraph{Source and scope.}
The motions come from the AMS synthetic balance-motion dataset~\citep{ams2026}: 10,000 single-leg balance motions produced by AMS's trajectory-optimization pipeline, of which we discard 186 clips with ground penetration, leaving 9,814 valid clips. Because these lie on essentially one single-support motion manifold, their training information is highly redundant, so we draw a compact, class-balanced random subsample of 900 (seed-fixed, uniform over the pose classes and support side) that gives controlled, uniform coverage of the pose space at low compute. Every clip is a fixed-length 4.98\,s trajectory (249 frames at 50\,Hz) built around a single-support balance phase, transitioning from double to single and back to double support; the middle single-support window is the phase our benchmark evaluates.

\paragraph{Two orthogonal pose axes.}
We stratify by two pose descriptors, each a per-clip mean over the full trajectory: support-leg squat depth (pelvis height) and swing-foot absolute height. We use the swing foot's absolute height rather than its height relative to the pelvis because the absolute height is essentially uncorrelated with pelvis height ($+0.03$), giving two orthogonal axes, whereas the relative height is strongly anti-correlated ($-0.53$): a deep squat mechanically inflates the relative lift, so a relative-height grid would conflate the two axes and starve corners such as ``deep squat, low lift.'' Each axis is cut into three bins, giving a $3\times3$ grid of nine classes. The pelvis-height edges are $0.38/0.51/0.65/0.78$\,m and the swing-foot edges $0.09/0.28/0.48/0.67$\,m.

\paragraph{Balanced coverage and leak-free split.}
Each of the nine classes holds exactly 100 clips, balanced 50/50 between left- and right-support (450/450 overall). We split 720 train / 180 held-out (80/20 per class), and further partition the 180 into a 90-clip validation set (for checkpoint selection) and a 90-clip test set (reported), both balanced across the nine classes and both support sides (seed 42). Because model selection touches only validation and reporting only test, the reported generalization neither peeks at the test set nor overfits the training distribution.

\section{Metric Definitions and Thresholds}\label{app:metrics-def}

\paragraph{Recovery sub-modes.}
A Marginal success (\cref{sec:protocol}) is reached by hop-recovery (the stance foot briefly leaves and returns), by touchdown-recovery (the swing foot lands), or by both. We report the two sub-modes separately because each independently marks a departure from a clean single-leg hold: a hop means the planted foot did not stay planted, and a touchdown means the robot is stepping to stay up rather than holding a single-leg stance at all. Empirically the two almost always fire together: 88.5\% of Marginal clips show both a touchdown and a hop, and none a static drift. A Marginal recovery is therefore typically a coupled step-and-hop shuffle rather than either mode alone. A held leg that drifts outside the 0.5\,m tracking gate would also score Marginal rather than Failure; empirically this case does not occur.

\paragraph{Continuous diagnostics.}
All balance-axis quantities are computed on the true whole-body \ac{com} (the simulator's mass-weighted per-link center of mass), not on the link-origin proxy the actor observes; the proxy-to-\ac{com} offset is quantified in \cref{tab:proxy-err}. The \ac{mos} is the primary continuous measure; we also report the actual-\ac{com} margin $m_c$, the capture-point out-of-support duration, HuB's 12-keypoint tracking errors $\Epos/\Evel/\Eacc$, control jerk, and \ac{ttf}. Support-foot slippage is the continuous, sub-threshold version of the same base reconfiguration that the hop and touchdown gates detect; we report it as a contact-quality diagnostic rather than a gate because it has no principled zero: even a cleanly planted foot slides under sensor and contact noise, so any cutoff would be arbitrary. One substitution is forced by running heterogeneous external policies at the command interface: not every method exposes its raw action, so smoothness is measured as the \ac{rms} of the second difference of the true joint velocities, a smoothness proxy rather than physical jerk, comparable across all methods.

\paragraph{Thresholds and windows.}
The evaluation fixes these constants for every method. A foot counts as in contact when its vertical ground-reaction force exceeds $0.05\,mg$. A trial is scored as a fall when the base height drops below half its nominal standing value or the base tilt exceeds 1\,rad. The swing-foot touchdown check trims 0.2\,s from each end of the single-support window, so a foot lifted slightly early or set down slightly late at the window edges is not counted as a touchdown. Keypoint tracking uses HuB's 0.5\,m average 12-keypoint gate. The clean protocol is deterministic ($K{=}1$); the noisy protocol averages ten fixed seeds ($K{=}10$, seeds 0--9).

\section{Full Continuous Metrics}\label{app:metrics}

\Cref{tab:metrics-full} reports the complete continuous suite for all nine methods.

\begin{table}[t]
    \caption{\textbf{Full continuous metrics.} Held-out test set, clean and deterministic, behind the balance summary of \cref{tab:main}. \acs{mos} and $m_c$ are the true-\acs{com} capture-point and actual-\acs{com} margins to the support boundary (\acs{ml}\,/\,\acs{ap}; less negative is better; per-motion minimum over the single-support window, averaged over the 90 clips). \acs{xcom}-out: seconds the capture point is outside support (single-support window ${\approx}3.3$\,s). $\Epos/\Evel/\Eacc$: 12-keypoint tracking error (mm / mm per frame / mm per frame$^2$). Slip: support-foot slippage (mm/s). TrackFail: \% of clips with mean keypoint error above 0.5\,m. Jerk: joint-velocity second-difference \acs{rms} (a smoothness probe, not divided by $dt$). Hop: mean support-foot hop count. \acs{ttf}: time-to-fall (s). All rollouts use MuJoCo; four methods use the shared DDC plant and four use method-specific, release-derived setups. Continuous values are descriptive rather than a one-plant ranking. (\cref{app:deploy}). Best in \textbf{bold}.}
    \label{tab:metrics-full}
    \centering
    \small
    \setlength{\tabcolsep}{3pt}
    \begin{tabular}{@{}l cccccc@{}}
        \toprule
        Method & \acs{mos}$_{\text{\acs{ml}/\acs{ap}}}$ & $m_{c,\text{\acs{ml}/\acs{ap}}}$ & \acs{xcom}-out & $\Epos$ & $\Evel$ & $\Eacc$ \\
        \midrule
        \textbf{\method} & \textbf{$-$.025\,/\,$+$.048} & \textbf{$-$.036\,/\,$+$.055} & \textbf{0.09} & \textbf{98} & \textbf{3.5} & \textbf{0.6} \\
        \midrule
        SONIC & $-$.281\,/\,$-$.209 & $-$.143\,/\,$-$.086 & 2.92 & 639 & 11.2 & 2.9 \\
        Humanoid-GPT & $-$.218\,/\,$-$.092 & $-$.153\,/\,$-$.035 & 3.08 & 289 & 7.0 & 1.9 \\
        HoloMotion & $-$.304\,/\,$-$.174 & $-$.250\,/\,$-$.127 & 3.07 & 333 & 7.0 & 1.8 \\
        MOSAIC & $-$.376\,/\,$-$.270 & $-$.264\,/\,$-$.189 & 2.98 & 499 & 9.2 & 2.4 \\
        ProtoMotions & $-$.314\,/\,$-$.250 & $-$.232\,/\,$-$.150 & 3.06 & 654 & 10.8 & 3.0 \\
        TWIST & $-$.615\,/\,$-$.527 & $-$.409\,/\,$-$.321 & 3.24 & 918 & 18.8 & 6.7 \\
        GMT & $-$.660\,/\,$-$.460 & $-$.498\,/\,$-$.341 & 3.19 & 874 & 14.6 & 4.3 \\
        OmniXtreme & $-$.736\,/\,$-$.611 & $-$.574\,/\,$-$.483 & 3.25 & 1087 & 18.4 & 6.1 \\
        \midrule
        \addlinespace
        Method & Slip & TrackFail & Jerk & Hop & \acs{ttf} & \\
        \midrule
        \textbf{\method} & \textbf{4.3} & \textbf{1.1} & \textbf{0.23} & \textbf{0.1} & \textbf{5.0} & \\
        \midrule
        SONIC & 135 & 52 & 1.01 & 8.7 & 4.6 & \\
        Humanoid-GPT & 54 & 16 & 0.70 & 5.6 & 4.4 & \\
        HoloMotion & 84 & 26 & 0.67 & 7.2 & 4.2 & \\
        MOSAIC & 131 & 43 & 0.97 & 8.6 & 4.2 & \\
        ProtoMotions & 173 & 51 & 1.60 & 14.5 & 3.6 & \\
        TWIST & 290 & 84 & 2.50 & 14.6 & 3.8 & \\
        GMT & 304 & 89 & 1.53 & 12.7 & 2.8 & \\
        OmniXtreme & 378 & 96 & 2.59 & 10.5 & 2.2 & \\
        \bottomrule
    \end{tabular}
\end{table}

\section{Deployment and Harness Details}\label{app:deploy}

\paragraph{The deployed policy.}
We deploy the benchmark-selected checkpoint (\cref{sec:ckpt}) on a physical Unitree G1 (29 \acp{dof}). The same trained actor is exported to ONNX and run directly at 50\,Hz, the training and benchmark control rate: the policy on the robot is the policy evaluated in simulation.

\paragraph{On-board observation.}
Every robot-state feedback term the actor consumes is reconstructed on-board from the joint encoders and the \ac{imu}; the reference terms (support phase and the future window) are provided by the baked motion command, and the previous action is maintained internally by the controller. All observations are expressed in the base (pelvis) frame used in training; on our G1 we take the SDK's \ac{imu} attitude directly in this frame, with no waist conversion. The dynamic-\ac{com} observation (\cref{sec:obs}) is computed on hardware from the encoder joint positions and velocities, the \ac{imu} gyroscope, and projected gravity alone; the base's absolute position and linear velocity, which no on-board sensor measures directly, are held at zero in the state buffer, where the support-relative differencing cancels them exactly (\cref{app:cancel}). The support foot is chosen by the same gravity-aligned foot-height rule as in training (the lower foot supports; feet within 3\,cm count as double support), so the support switch has no train-to-deploy discontinuity.

\paragraph{Control and processing.}
The actor outputs a residual about the default pose that becomes a joint-position target for the robot's on-board \ac{pd} controller. Deployment adds only minimal, standard processing: numeric clipping on observations and actions, and a light low-pass filter on the waist encoders to suppress sensor noise.

\paragraph{Harness per method.}
ProtoMotions, MOSAIC, SONIC, and HoloMotion reuse the \method G1 plant (the same MuJoCo model and 50\,Hz control loop; each supplies its own \ac{pd} gains and action scaling); GMT, TWIST, OmniXtreme, and Humanoid-GPT run in their own validated deployment harness. The split follows each release: a method that ships a runnable G1 deployment loop runs in it, and a method released as weights only is hosted on our plant through a thin adapter that reconstructs its documented observations (HoloMotion's own observation kernel is imported unchanged). The 0.5\,m keypoint gate in the Perfect definition binds only once a method sustains single-leg support; as no baseline does, it never changes a baseline's tier. OmniXtreme is additionally a stochastic flow policy that samples fresh initial noise per run: its Perfect tier is seed-invariant (0/90), but its Marginal/Failure split and continuous margins vary run-to-run, so we report a fixed default-seed sample.

\end{document}